\RequirePackage{fix-cm}
\documentclass{svjour3}                     
\smartqed  
\usepackage{graphicx}
\graphicspath{ {./figures/} }
\DeclareGraphicsExtensions{.pdf,.jpeg,.png,.emf,.jpg}
\usepackage{multirow}
\usepackage{mathptmx} 
\usepackage{times} 
\usepackage{amsmath} 
\usepackage{amssymb}  

\begin{document}

	\title{Leveraging Vision and Kinematics Data to Improve Realism of Biomechanic Soft-tissue Simulation for Robotic Surgery}
	
	\author{Jie~Ying~Wu \and Peter~Kazanzides \and Mathias~Unberath 
	}
	
	\institute{J. Wu, P. Kazanzides, and M. Unberath are with the Department of Computer Science, Johns Hopkins University, Baltimore, MD 21218 USA (email: {jieying@jhu.edu; pkaz@jhu.edu; unberath@jhu.edu})}
	
	\date{Received: date / Accepted: date}
		
	\maketitle

	\begin{abstract}
	
	 \section*{}
	 \textbf{Purpose} Surgical simulations play an increasingly important role in surgeon education and developing algorithms that enable robots to perform surgical subtasks. To model anatomy, Finite Element Method (FEM) simulations have been held as the gold standard for calculating accurate soft-tissue deformation. Unfortunately, their accuracy is highly dependent on the simulation parameters, which can be difficult to obtain.\\
	 \textbf{Methods} In this work, we investigate how live data acquired during any robotic endoscopic surgical procedure may be used to correct for inaccurate FEM simulation results. Since FEMs are calculated from initial parameters and cannot directly incorporate observations, we propose to add a correction factor that accounts for the discrepancy between simulation and observations. We train a network to predict this correction factor.\\
	 \textbf{Results} To evaluate our method, we use an open-source da Vinci Surgical System to probe a soft-tissue phantom and replay the interaction in simulation. We train the network to correct for the difference between the predicted mesh position and the measured point cloud. This results in 15-30\% improvement in the mean distance, demonstrating the effectiveness of our approach across a large range of simulation parameters.\\
	 \textbf{Conclusion} We show a first step towards a framework that synergistically combines the benefits of model-based simulation and real-time observations. It corrects discrepancies between simulation and the scene that results from inaccurate modeling parameters. This can provide a more accurate simulation environment for surgeons and better data with which to train algorithms. \\
		
	\end{abstract}
	
	\section{INTRODUCTION}
	
	Robotic surgery has changed the way many surgeries are performed. 
	The da Vinci Surgical System\textregistered~ (Intuitive Surgical Inc., Sunnyvale, CA) is the most successful example with more than 4000 systems installed around the world. 
	Not only do robots allow physicians to perform more complex surgeries, they also open the possibility for machine learning algorithms provide aid, such as automating 
	surgical subtasks. Obtaining data to train these algorithms can be difficult, including but not limited to concerns regarding patient privacy; however, 
	before operating on patients, doctors must practice on simulators. This may provide a valuable source of data. 
	Unfortunately, many simulators have rudimentary physics and cannot accurately model large deformations. Thus, these simulators test surgeons on simplified tasks to train agility rather than on a full surgery. 
	While surgeons are adept at generalizing from these tasks to the clinic, algorithms are limited by the data they are provided.  
	More accurate simulators are necessary to create realistic tasks, which can both 
	benefit the surgeon and provide high quality data to train the robot to intelligently aid physicians. 
	
	High precision is required for tissue simulation for medical use. 
	FEM is the current gold standard for simulating deformation in soft-tissue; however, its use in patient 
	modeling is limited by inaccuracies in parameter estimation and its computational complexity.
	Accurate material parameters are integral to accurate FEM simulations, though boundary conditions and geometric model also play an important role \cite{pfeiffer2019learning}. These 
	are often difficult to measure in a phantom, and impossible to obtain in a clinical setting.
	Inaccurate models in the FEM simulations result in inaccurate predictions of tissue behavior.
	Given these limitations inherent in any model, we propose a neural network that can provide corrections to an FEM model in the presence of large deformations and inaccurate material parameters.
	The network is self-supervised as it can be trained from data available in any robotic surgery: endoscope video and robot kinematics. 
	Recent works have recovered the 3D surface during surgery from the endoscope video \cite{liu2019dense}, which provides the ground truth for our network.
	
	This is a first step of a framework that could be capable of life-long learning and adapt models to patients 
	throughout a surgery. This can provide better feedback for users and serve as 
	a basis for generating datasets for data-driven methods. Current surgical subtask automation 
	research on soft-tissue manipulation is limited to simple 2D phantoms~\cite{shin2019autonomous} or path planning algorithms with no tissue interaction~\cite{richter2019open}. The proposed system was implemented on the
	first generation da Vinci Surgical System with the open source controllers of the da Vinci Research Kit (dVRK)~\cite{kazanzides2014open}. We summarize our contribution as follows:

	\begin{itemize}
		\item One of the first biomechanically accurate soft-tissue simulators that is fully integrated with the dVRK framework 
		\item A data-driven approach to improve FEM simulations of soft-tissue deformation based on robotics vision and kinematics data
	\end{itemize}
	
	We understand our approach to be a first step in a framework that synergistically combines the benefits of model-based simulation and real-time observations.
	
	\section{PREVIOUS WORK}
	
	We first review the simulation platforms that exist for the dVRK. Second, we review the most relevant learning algorithms that have been proposed to improve simulations. 
	
	\paragraph{Simulators}

	Many existing works on simulators for the da Vinci Research Kit have focused on the robot kinematics. Gondokaryono et 
	al. uses Gazebo to support environment objects like placing a camera but does not 
	actually interact with any object ~\cite{gondokaryono2019approach}. This work was extended to include interactions by Munawar et al., which introduced a new simulation environment~\cite{munawar2019a}. 
	The simulation environment uses a new robot description format and supports soft bodies. The aim here is rapidly prototyping complex environments for surgeon training rather than accurate physics.
	Fontanelli et al. uses	V-REP to create a rigid body simulation of the da Vinci but does not 
	support soft-tissue interaction due to V-REP's limited physics backend~\cite{fontanelli2018v}. 
	They discuss the desirability of a soft-tissue simulator and suggested using the SOFA framework as the physics simulator.
	The SOFA framework~\cite{allard2007sofa} focuses on simulation for medical purposes and has been validated on a 
	variety of tasks from surgical training to guidance~\cite{talbot2015surgery}. In this work, since we are interested in modeling the soft-tissue interaction, we chose SOFA as the environment.

	\paragraph{Learning to Correct Simulations}

	While FEMs are the standard today for accurate, deformable tissue 
	simulation, other models have been proposed and the problem of parameter estimation is common. We refer readers 
	to the survey by Zhang et al.~\cite{zhang2017deformable} for details on FEM simulation and a comparison to other deformable models.  
	Bianchi et al. learned parameters of simulations by approximating soft tissue as 
	spring-mass models~\cite{bianchi2004simultaneous}. They found that a system with homogeneous stiffness is 
	insufficient to model the complexity of soft-tissue behavior. 

	More recently, researchers have begun training deep learning models to predict deformations from FEM models. Morooka et al. 
	trains a network to predict the deformations of an FEM, where the input is a force vector and 
	contact point, and the output is every point of the deformed mesh~\cite{morooka2008real}. The size of the simulation 
	object must be strictly reduced since their network is fully-connected and thus the network size is scales poorly with input and output size. 
	They use principal component analysis (PCA) to keep their network size tractable. Although 
	their method could be trained end-to-end by using more recent architectures such as autoencoders, these are 
	currently shown to provide limited advantage over PCA to model mesh deformations~\cite{roewer2018towards}. 	
	Meister et al. presents an alternative approach to use a fully-connected network to predict the solution 
	of the Total Lagrangian Explicit Dynamics needed for FEM for each vertex of the mesh. They show that this is stable for 
	larger timesteps than those at which FEM is stable and could be used to speed up deformation~\cite{meister2020deep}.

	Other works have replaced the fully-connected layers with convolutional layers to process larger meshes. 
	Mendizabal et al. use a 3D-UNet architecture~\cite{cciccek20163d} to learn the deformation of 
	a mesh given forces represented by a 3D grid~\cite{mendizabal2020simulation}. They apply simulated force to a known mesh 
	and generate the desired deformation using FEM. Their network aims to predict the FEM deformation. Pfeiffer et al. use a similar U-Net 
	architecture to fill in the deformation of the entire organ given the deformation on a
	partial surface~\cite{pfeiffer2019learning}. They show that their synthetically trained network 
	can generalize to real cases, both in phantom and human CT data, with different geometry. Our work is complementary in that we predict the surface deformation given a robot position for a 
	mesh. Both these works apply a force on FEM and measure the steady state response rather than capture the collision dynamics between objects. 
	
	\section{METHOD}
\begin{figure}[h!]
		\center
		\includegraphics[width=\columnwidth]{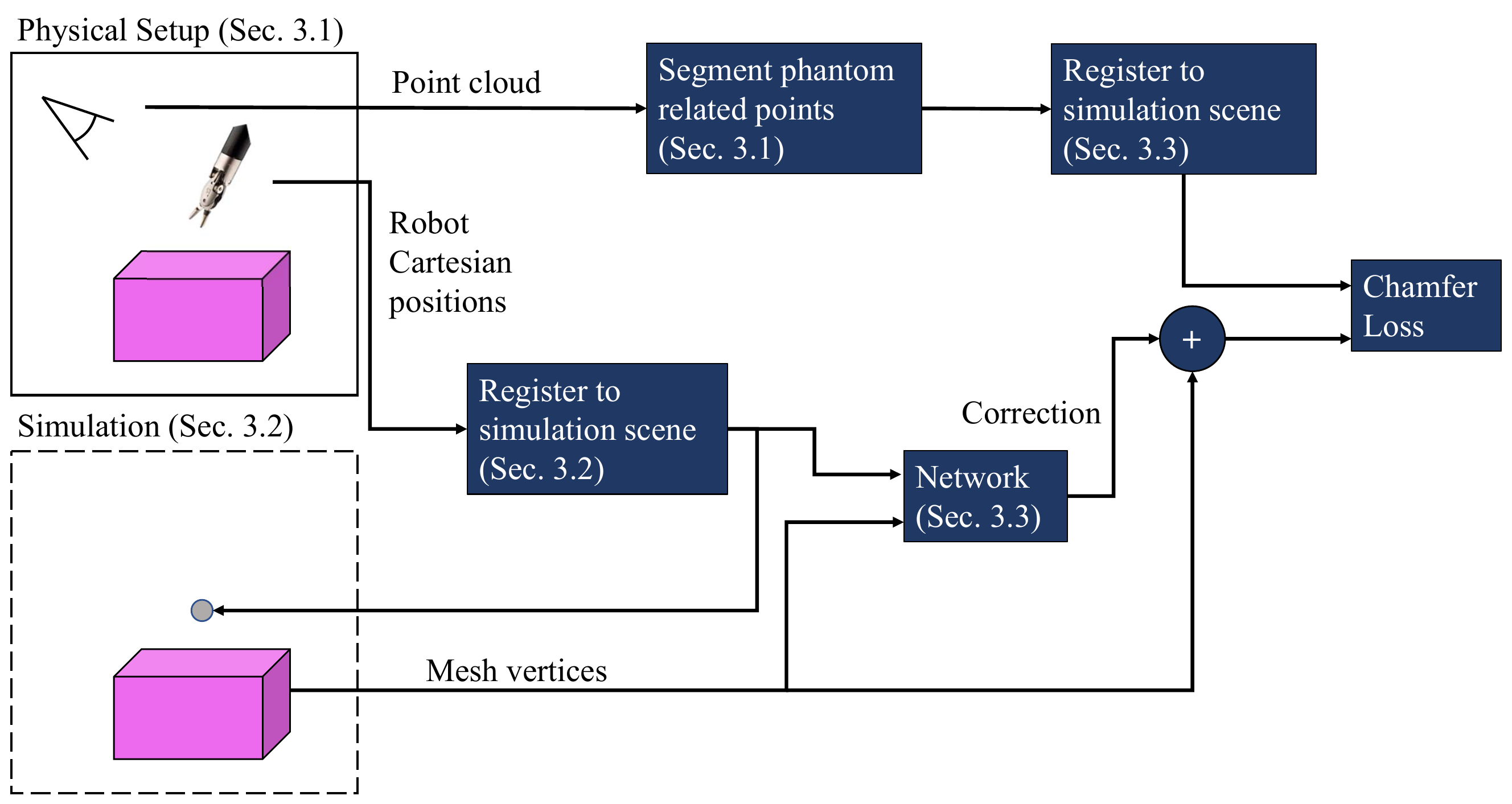}
		\caption{Overview of the proposed system. The left blocks show the physical setup and its corresponding simulation scene. The system captures the robot Cartesian positions as it is moved and registers them to the simulation scene, where it is replayed. The mesh vertices from the simulation are read out and fed into a network which predicts a correction factor. It is trained by comparing the simulated mesh plus the correction factor with the point cloud captured from the physical set up.}
		\label{fig:block_diagram}
	\end{figure}

	Our goal is to learn correction factors for an FEM simulation by comparing the simulation results to a physical setup. To that end, there are three main components to our setup. First, 
	we create a physical phantom and use the da Vinci patient-side manipulator (PSM) to palpate the phantom and capture the RGBD video and robot kinematics.
	Second, we create the same scene in simulation and replay the interaction to simulate the shape of the phantom as it deforms from the robot interactions. 
	Lastly, we compare the simulation mesh vertex positions and the measured position of the physical phantom and train a network to correct the simulated positions. Fig.~\ref{fig:block_diagram} shows an overview of the setup.

	\subsection{Physical Setup}
\begin{figure}[h!]
		\center
		\includegraphics[width=0.45\columnwidth]{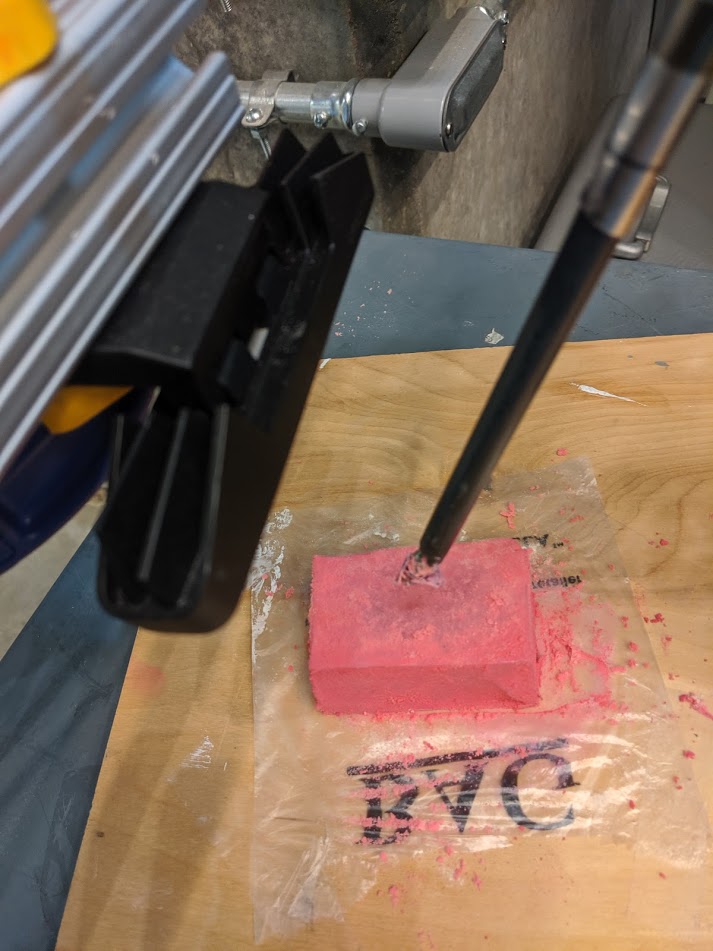}
		\caption{Deformable phantom setup. The RGBD camera is mounted above the workspace on the left. The phantom is coated with spray paint and potato starch to improve its visibility to the depth camera.}
		\label{fig:deformable_phantom}
	\end{figure}

	Fig.~\ref{fig:deformable_phantom} shows the phantom setup. The robot instrument is positioned to the top right and the depth camera is on the left, mounted above the workspace. We move the robot 
	instrument to interact with the phantom while capturing its Cartesian positions through ROS~\cite{quigley2009ros}. The camera measures the deformation of the phantom. Interactions consist of probes to the top and sides of the phantom. 	Each frame of the depth camera is read out as a point cloud. We use the Point Cloud Library (PCL)~\cite{rusu20113d} 
	to remove points from the instrument and the table, as well as outliers so that the only points that remain are 
	the ones from the phantom. Then we subsample the points to about 45k points, or about 16.5 points per mm$^2$.

	\subsection{Simulation in SOFA}
	We create the surface mesh of the phantom in Solidworks (Dassault Syst\`{e}mes,
	V\'{e}lizy-Villacoublay, France) and fill it with Gmsh~\cite{geuzaine2009gmsh} to create a solid, tetrahedral mesh. 
	SOFA provides different templates for modeling objects; here we focus on two of them. With the `Rigid3d' template, 
	objects have 7 parameters (3 translational and 4 for rotation quaternion) while 
	with the `Vec3d' template, each vertex of the object's mesh can be individually set. We use Vec3d for the phantom and Rigid3d for the instrument. 

\begin{figure}[h!]
		\center
		\includegraphics[width=0.8\columnwidth]{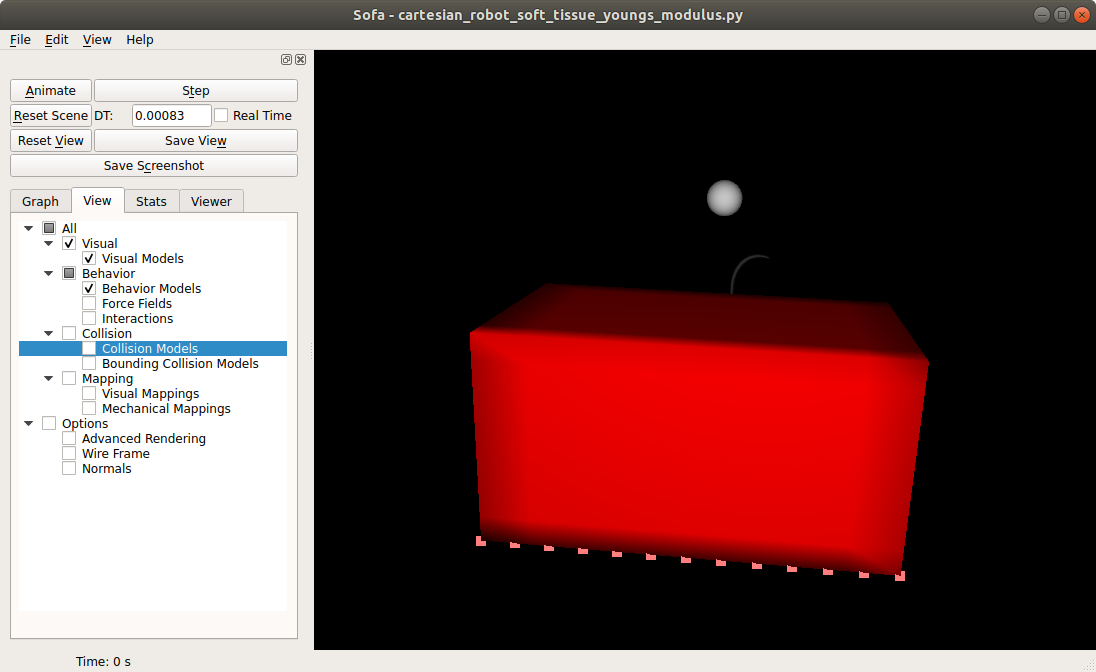}
		\caption{Simulated phantom. The gray sphere is the end-effector. The red block is the deformable phantom, with the pink cubes showing where the phantom mesh is fixed.}
		\label{fig:simulated_phantom}
	\end{figure}

	While we implemented the robot in SOFA, computing the constrained movements was too computationally expensive and not necessary for learning the mesh deformation. Instead, to simplify computation, we only simulate the end-effector pose and approximate it as a sphere of radius 5 mm. The sphere is moved based on the Cartesian position of the end-effector at each step. 
	Additionally, to avoid simulating computationally expensive friction interaction with the table, we suspend the phantom by fixing the bottom vertices in 
	space rather than placing it on a plane. The fixed vertices are marked by pink cube overlays in Fig.~\ref{fig:simulated_phantom}.
	Gravity is currently not simulated. 
	We use a parameter search to find the best material parameters for the phantom. The results are shown in section~\ref{sec:parameters}.
	
	We calibrate the real robot to the simulator reference frame. We capture
	a calibration sequence where we move the robot to touch each of the four corners of the phantom 
	and extract the corner location from the robot and then perform a rigid registration to the corresponding points of the simulated phantom. 
	The simulated robot replays the calibrated actions and the simulation saves the positions of the vertices of the phantom mesh at each timestep that has a corresponding point cloud label. 
	At each timestep, the robot end-effector position is updated based on the captured kinematics from the interactions of the physical robot. 
	To ensure the FEM simulations run stably, the simulation timestep is much smaller than the rate at which the robot kinematics 
	is captured. We linearly interpolate the robot end-effector position for simulation timesteps where we do not have a corresponding label.
	
	\subsection{Estimating Correction Factors}
	
	We train the network using the pre-calculated FEM-simulated phantom mesh vertex positions as input. The output 
	is a correction factor for the vertex positions to match the real, measured positions.

	The vertices of the phantom mesh are represented by a 3D matrix, where each element has three values for the 3D position of that vertex. 
	We follow the architecture proposed by previous work and use a 3D-UNet~\cite{cciccek20163d} to process the mesh. 
	The network contains 3 blocks, each with 2 convolutional layers with kernel size of 3 and feature 
	size of 64, 128, and 256 respectively. There is a Max-pooling layer between each block with kernel size of 2 to downsample the mesh. We keep the default activation function, ReLU. 
	The kinematics information is inserted at the bottom of the network, where the spatial 
	information is the most condensed. The robot pose is concatenated as additional features to each of the voxels. This is inspired by the work of Finn and Levine~\cite{finn2017deep} in robot-motion planning where the action is inserted in the 
	middle of the video processing pipeline so the first layers focuses solely on images. We add 4
	convolutional layers, also with kernel size 3, to incorporate the kinematics information. Since 
	the kinematic information is added as features, we also use them to reduce the feature size back to 256 so that the feature sizes match when concatenated with previous layers for the decoder. 
	
	Here, we want the network to extract global mesh information before changing the portion of the mesh with which the robot is interacting. The end-effector position is concatenated to each of the 
	voxels before the decoding layers.  
	The network learns which part of the mesh the end-effector affects and produces a correction step for the displacement of each vertex so the output is the same size as the input. The correction is added to the mesh and we compare this to the point cloud captured 
	by a depth camera for ground truth displacements. 
	
	To register from the camera to the simulation, we extract the point cloud of the phantom without any robot interaction. 
	We find the transformation between the phantom and that point cloud using iterative closest point (ICP)~\cite{besl1992method} as implemented in PCL. 
	Since the top of the block is flat and featureless, we manually initialize the registration so that the algorithm registers to the correct pose. 
	The depth camera only provides labels for the top layer of the phantom. Therefore, we only allow the correction to be applied to the top layer of the mesh. We also test a 2D version of UNet where only the top layer of the mesh is passed to the network.	
	
	\begin{figure}[h!]
		\center
		\includegraphics[width=0.8\columnwidth]{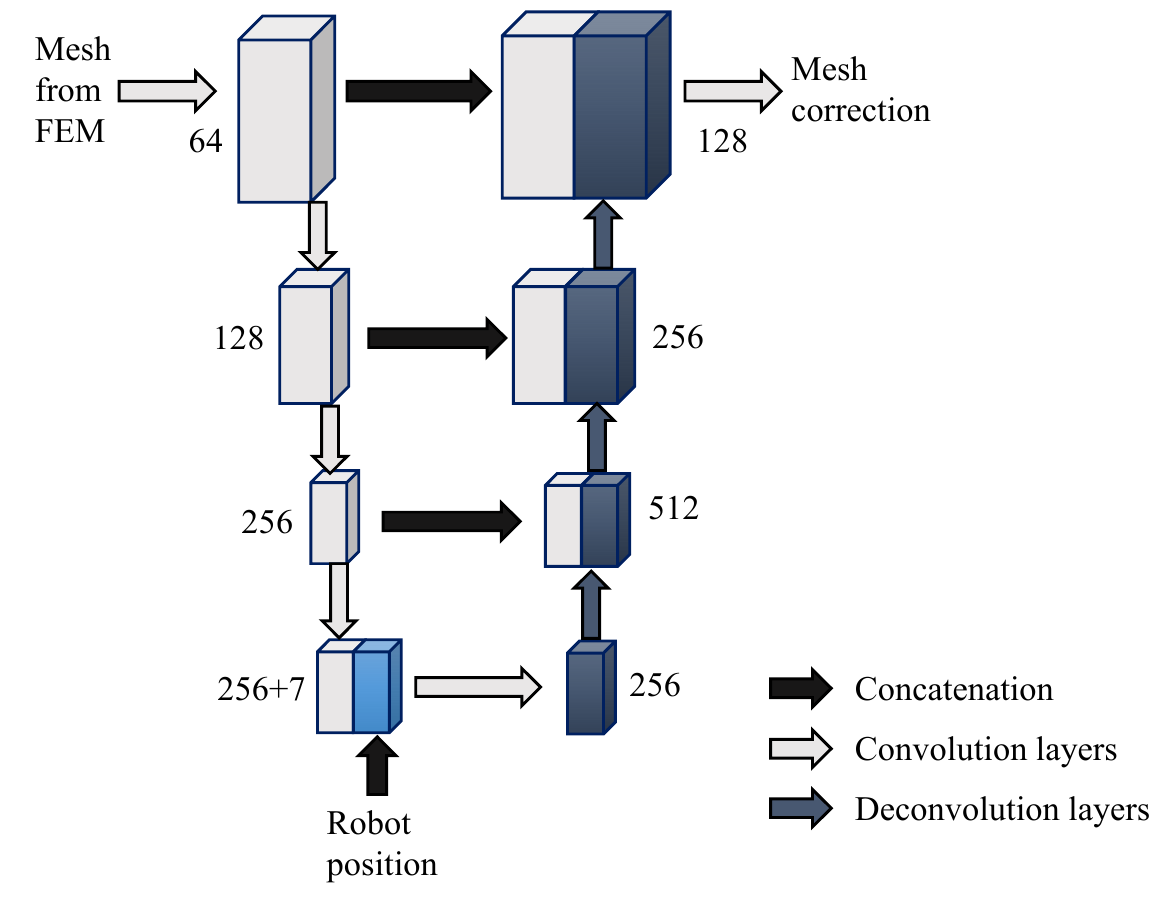}
		\label{fig:network}
		\caption{The network architecture for calculating the mesh correction. We modify the 3D-UNet~\cite{cciccek20163d} to 
			insert the kinematics at the bottom, as shown by the light blue block. The robot position is concatenated to 
			every voxel of the mesh as a feature. Each block features two convolutional layers and one Max-pooling. The numbers by each block indicate the number of features. All convolutions are with kernel size 3 and padding 1. }
	\end{figure}
	
	At each simulation step, we pass the position of the phantom and the robot position as input to a neural network. 
	The network predicts a correction factor and updates its weight by comparing the corrected mesh and actual phantom position. 
	To prevent vertices from crossing, the maximum correction in each dimension is scaled to be 
	half of the voxel length in each direction. 
	The training loss is the Chamfer distance, or the mean shortest distance between one point on a 
	point cloud to another point cloud. In this case, we represent our mesh as a point cloud of its vertices. We 
	initially trained with Hausdorff distance but since it is sensitive to outliers, the training 
	did not converge well. Using the network to calculate a displacement map rather than the positions 
	directly avoids the need for formulating a regularization term in the loss function to prevent the 
	vertex positions from crossing over each other. This also gives us an additional factor to control 
	how much we want to rely on the simulation. If we trust our FEM more, we can set the maximal 
	displacement to be lower, or higher if we are more uncertain about the FEM parameters.  
	
	During test time, at the end of each timestep, the simulator passes the position of the phantom 
	mesh to the trained network to get the correction. In the 2D network case, we only pass in the top layer of the mesh to the network. 
	In our training, we noticed that since our ground truth can only constrain the top layer of the mesh, only those values are valid and are used to update the 
	simulation. The other nodes stay as the simulator calculates them. 
	
	To compare the mesh to the point cloud, we super-sample the mesh by a factor of 3 in all directions using 
	linear interpolation. The mesh must be fairly sparse due to computation power constraints for the FEM simulation 
	but this leads to inefficient learning for the network as the loss is not smooth. 
	This does not change the form of the mesh but provides more points for 
	the point cloud to point cloud matching. Using this point cloud loss avoids the more computationally 
	expensive point-to-surface matching. We only calculate loss from the direction of the depth camera output to the mesh since there may be no correspondences in the other direction due to occlusion from the surgical instrument.

\section{EXPERIMENTS}	
	
	\subsection{Soft-tissue Phantom}
	The phantom was created by pouring liquid plastic into a mold with some amount of 
	hardener and softeners from M-F Manufacturing (Fort Worth, TX). After the phantom has 
	set, we cut out a rectangular block.
	After trimming, the phantom's dimensions are $68.7 \times 35.8 \times 39.3$ mm and it weighs 104.01 g. 
	We do not measure the stiffness of the phantom but perform a parameter search for its material parameters. 
	 
	We move a PSM directly, without teleoperation, to interact with the phantom while recording its kinematics and
	measuring the deformation of the phantom using an Intel Realsense SR300 camera (Intel, Santa Clara, CA). Intel reports the accuracy
	of the depth measurement to be 2 mm. Since the plastic is translucent after 
	setting, we coat the phantom in spray paint and potato starch so that it can be measured by the 
	depth camera. 
	
	In the first video, we capture data for calibration. We manipulate the robot so that its end-effector successively touches each of the phantom's corners 
	to calibrate between the physical setup and the simulated scene. During the interactions, robot position is captured at 1 
	kHz while the depth image is captured at 30 Hz. We subsample the robot positions to match that of 
	the depth images in time. After calibrating our setup, we capture 12 more interactions. 
	This results in around 14 minutes, or 25k frames, of video. Of those  
	interactions, the first one is 2 minutes and the rest are around 1 minute. The 2 minutes, 3794 frames, clip is used as 
	the validation set. The next two one-minute segments, 4417 frames, form the test set.
	
	\subsection{Simulation Scene}
	
	The model in SOFA is represented as a solid, tetrahedral mesh of size $13\times5\times$5. Even with the small mesh, it takes on the order of hours
	to simulate a minute of interactions. This may be improved by multi-threading but as that does not improve throughput over running multiple 
	instances, it was not implemented for this work. A finer surface mesh 
	is attached to that for smoother visualization. The da Vinci instrument end-effector is approximated as a sphere with comparatively large mass of 1000 g. 
	We set the minimum contact distance to be 0.5 mm and use Euler implicit solver with Rayleigh stiffness and mass both set to 0.1. 
	
	\subsection{Material Parameter Search}
	\label{sec:parameters}
	The main parameters of soft tissue can be considered to be its Young's Modulus and Poisson's Ratio. Uncertainty 
	in the parameters come from the unknown mixture of softeners and hardeners during construction of the phantom as well as its age.
	To limit the search space and since the two parameters are interdependent, we focus our parameter 
	search on the Young's Modulus. Values for the Poisson's Ratio are more consistent in literature to be in the 0.4-0.5 
	range, whereas estimates of the Young's Modulus vary anywhere in the range of 1e-4 to 1e6 depending on the specific 
	construction of the phantom. We set the Poisson's ratio based on previous work~\cite{lee2013novel}.
	
	To perform a grid search for the optimal Young's Modulus, we choose one interaction sample from the training set, about 1 min in duration, and simulate it with different Young's Modulus. Since FEMs are computationally 
	expensive to run, we separate the test in two stages. First, we sample the Young's Modulus at every factor of 10 from 1e1 to 1e6 to explore a broad range of values. We write out the simulated 
	phantom positions, super-sample the mesh by a factor of 3, and compare the vertex positions to the measured point cloud. The average distance from a point on the depth camera point cloud to the simulated mesh are reported in Tab.~\ref{tab:parameter_coarse}. 
	
	\begin{table}[h!]
		\begin{center}
			\caption{A one-minute interaction with the phantom is simulated with each of the Young's Modulus and the average distance in mm from measured point cloud to the mesh vertices is reported.}
			\label{tab:parameter_coarse}
			\begin{tabular}{ c | c | c | c | c | c }
				1e1 & 1e2 & 1e3 & 1e4 & 1e5 & 1e6\\ 
				\hline
				8.4981 & 6.0760 & 5.5417 & 5.2632 & 5.1847 & 9.3356
			\end{tabular}
		\end{center}
	\end{table}
	
	Next, we pick the optimal range and explore that more finely, in this case around 1e3 to 1e5. We test each interval at every 2.5e3 and 2.5e4 respectively. Results are shown in Tab.~\ref{tab:parameter_fine}. 
	
	\begin{table}[h!]
		\begin{center}
			\caption{A one-minute interaction with the phantom is simulated with each of the Young's Modulus and the average distance in mm from measured point cloud to the mesh vertices is reported. The simulation with Young's Modulus of 5e4 did not converge.}
			\label{tab:parameter_fine}
			\begin{tabular}{ c | c | c | c | c | c }
				2.5e3 & 5e3 & 7.5e3 & 2.5e4 & 5e4 & 7.5e4\\ 
				\hline
				5.2724 & 4.9041 & 5.5025 & 5.2202 & N/A & 5.4565
			\end{tabular}
		\end{center}
	\end{table}
	
	We observe that Young's Modulus optimization is not convex, potentially due to
	measurement error since point clouds are generally noisy, and that a small change in parameter can lead to a result that did not 
	converge. This reinforces the need to find a way to learn the optimal parameters rather than search exhaustively. 5e3 was 
	selected to be the optimal Young's Modulus. We simulate all the interactions using 5e3 to generate the input to our network. 
	
\section{RESULTS}

\subsection{Network Correction}
	We train the network on 9 min of the captured video samples, or 15872 samples, until convergence. Convergence is measured by loss on the 2 min of validation data flattening or going 
	up. We test on the remaining two sequences of about 2 min. The mesh is refined before calculating the Chamfer loss to provide a smoother loss for the network to train on. We compare 
	the performance using 2D and 3D network on two sequences as shown in table~\ref{tab:results_YM}. 
	One interesting use case for the network is when you have sub-optimal FEM parameters. Since the 
	parameter search is time-consuming, we test if the network can correct for non-optimal 
	parameters. We run the simulations with the same setup, but set the Young's Modulus to 1e4 and 
	1e1. Then we train and test the network using the same data split as before.

	\begin{table}[h!]
		\caption{The loss from FEM directly and after correction using 2D and 3D networks. The loss is the average distance between each point on the point cloud to the closest super-sampled mesh vertex in mm.}
		\label{tab:results_YM}
		\begin{center}
			\begin{tabular}{ c | c | c | c | c | c | c }
				& & No network & 2D network & Improvement & 3D network & Improvement\\
				\hline
				\multirow{2}{*}{5e3} & Sequence 1 & 5.4627 & 4.3025  & 21.2\% & 4.3234 & 20.9\% \\    
				\cline{2-7} 
				& Sequence 2 & 5.1170 & 4.3450 & 15.1\% & 4.6269 &  9.6\%\\
				\hline
				\multirow{2}{*}{1e4} & Sequence 1 & 5.2333 & 4.2323 & 19.1 \% & 4.6446 & 11.2\%\\    
				\cline{2-7} 
				& Sequence 2 & 5.2734 & 4.3746 & 18.4\% & 4.9828 & 5.5\%\\    
				\hline
				\multirow{2}{*}{1e1} & Sequence 1 & 8.2322 & 5.6235 & 31.7\% & 5.9614 & 27.6\% \\
				\cline{2-7}
				& Sequence 2 & 9.2899 & 6.4518 & 30.6\% & 6.7030 & 27.8\%\\   
			\end{tabular}
		\end{center}
	\end{table}

\section{DISCUSSION}

	We have shown a first step towards a framework to train networks to learn the behavior of soft-tissue directly from 
	robotic surgery data. This approach combines model-based simulations with real-time observations to correct for inaccurate simulation parameters. 
	Using the surface deformation and a robot position, we trained the network to improve 
	FEM results by 15-30\% over a range of simulation parameters. This network
	can be used to improve FEM results where we do not have reliable material parameter and implicitly adjust for 
	boundary conditions and other factors that are often intractable to model. We show that using the proposed correction 
	factor, even starting from a Young's Modulus 5e2 away from the optimal that we found, the network reduces the error to within 1.5 mm of the most accurate simulation without correction. 
	
	From our parameter search, we see that the best Young's Modulus found on one video sequence is not necessarily the best for other
	sequences. There are unmodeled forces that affect 
	deformation. Our proposed data-driven method avoids 
	the need to search the high-dimensional space for all these factors. 
	This observation reinforces the advantage of an observation-based correction. 
	Additionally, since the learning could be done online, this network can be updated during a procedure if the models do not match real observations.  
	The patient's soft-tissue characteristics may change during surgery and our method could be 
	adapted to do life-long learning, correcting for unmodeled changes.

	We expected the 3D-UNet to provide better correction than the 2D, since the network may take into account the position of the next-to-top vertices; however, the 2D network performed better.
	This may be due to not using enough training data as the 3D network has more parameters to train and not enough constraints for the non-top layers of the mesh. 
	The 3D network may benefit from integrating synthetic data and using the FEM to provide constraints for the unobservable vertices. This would provide ground truth for the entire mesh rather than only the surface. 

	More work is needed to show how well this method would work across different phantoms.	While any mesh can be interpolated to fit a regular grid, network architectures that work on meshes may show better performance on an arbitrarily shaped mesh. 
	Additionally, heterogeneous tissue is generally harder to model than homogeneous tissue and represent further opportunities for data-driven corrections. 
	Testing for how well networks trained on one model may generalize to another model of different stiffness or geometry would also be an interesting extension of this work. 
	
	We currently do not include gravity or the velocity of the end-effector, which may improve baseline simulation and the learned correction factors. We use default hyperparameters in 
	our network, and a careful search may improve results. While the network currently uses the FEM 
	output, it could in the future be trained directly without the FEM step. 
	This would be the natural extension of the existing literature that uses a network to predict FEM steps. Training on FEM results inherently limits their accuracies to be that of the 
	FEM and it would be interesting to incorporate data from physical interactions.

\section{CONCLUSION}
	In this work, we outline a framework to correct model-based simulation using data readily available in robotic 
	surgeries. We show a first step of such a system by implementing a network that predicts deformation 
	corrections on a soft-tissue phantom during robot interactions. This network can correct for inaccurately modeled parameters as we show improvements across a wide-range of Young's Modulus. 
	Other terms like friction and boundary conditions are rarely available in patient data so any model-based simulation would have a irreducible error.
	A data-driven correction factor could account for difficult-to-model errors. We envision that approaches similar to the one proposed here can be adapted for 
	patient use, performing life-long learning throughout a surgery as patient condition changes.

\section*{ACKNOWLEDGMENT}
\label{sec:acknowledgement}
Funding: This work is supported by National Science Foundation NRI 1637789. The Titan V used for this research was donated by the Nvidia Corporation. 
\\
Conflict of Interest: The authors declare that they have no conflict of interest.
\\
This article does not contain any studies with human participants or animals performed by any of the authors.

\bibliographystyle{spmpsci}
\bibliography{paper}	
	
\end{document}